\def\year{2020}\relax
\documentclass[letterpaper]{article} 
\usepackage{aaai20}  
\usepackage{times}  
\usepackage{helvet} 
\usepackage{courier}  
\usepackage[hyphens]{url}  
\usepackage{graphicx} 
\usepackage{graphicx}  
\usepackage{multirow}
\usepackage{graphicx}
\usepackage{xcolor}
\usepackage{algorithm}
\usepackage{algpseudocode}
\usepackage{amsmath}
\usepackage{booktabs}
\usepackage{graphicx}
\DeclareMathOperator*{\argmax}{argmax}   

\title{Enhanced Offensive Language Detection Through Data Augmentation}

\author{Ruibo Liu, Guangxuan Xu, Soroush Vosoughi\\ 
Department of Computer Science, Dartmouth College\\ 

Hanover, NH, USA\\
\{ruibo.liu.gr, guangxuan.xu.ug, soroush.vosoughi\}@dartmouth.edu 
}
 
\begin{document}
\linespread{0.93}
\maketitle
\frenchspacing
\begin{abstract}
Detecting offensive language on social media is an important task. The ICWSM-2020 Data Challenge Task 2 is aimed at identifying offensive content using a crowd-sourced dataset containing 100k labelled tweets. The dataset, however, suffers from class imbalance, where certain labels are extremely rare compared with other classes (e.g, the \textit{hateful} class is only $5\%$ of the data). In this work, we present \textbf{Dager} (\textbf{Da}ta Au\textbf{g}ment\textbf{er}), a generation-based data augmentation method, that improves the performance of classification on imbalanced and low-resource data such as the offensive language dataset. Dager extracts the lexical features of a given class, and uses these features to guide the generation of a conditional generator built on GPT-2. The generated text can then be added to the training set as augmentation data. We show that applying Dager can increase the F1 score of the data challenge by 11\% when we use 1\% of the whole dataset for training (using BERT for classification); moreover, the generated data also preserves the original labels very well. We test Dager on four different classifiers (BERT, CNN, Bi-LSTM with attention, and Transformer), observing universal improvement on the detection, indicating our method is effective and classifier-agnostic.
\end{abstract}

\section{Introduction}
Social media users get exposed to many kinds of abusive behavior such as hate speech, online bullying, and racist and sexist comments. Detecting and identifying abusive behavior can contribute to a more harmonious virtual environment. Given the huge amount of social media text produced every day, it is not realistic to rely on human annotators to manually filter offensive contents. Fortunately, recent years have witnessed substantial progress in computational methods to tackle the abusive detection problem, particularly on Twitter~\cite{davidson2017automated,zhang2019hate}.

In 2018, a crowd sourced dataset~\cite{founta2018large} was released to help model abusive behavior on Twitter. This dataset collects 100k tweets with four labels (\textit{normal}, \textit{spam}, \textit{abusive} and \textit{hateful}). The distribution of the labels for this dataset and two other related datasets is shown in Table~\ref{tab:data_distri}.

\begin{table}[H]
\centering
\scriptsize
\resizebox{0.47\textwidth}{!}{%
\begin{tabular}{|l|c|c|c|c|}
\hline
\multicolumn{1}{|c|}{\textbf{Datasets}} & \textit{normal} & \textit{spam} & \textit{abusive} & \textit{hateful} \\ \hline
\textbf{Founta et al. (2019)}                       & 53,851           & 14,030         & 27,150            & 4,965             \\ \hline
Davidson et al. (2017)                     & \multicolumn{2}{c|}{4,163}       & 19,190            & 1,430             \\ \hline
Waseem et al. (2016)                       & 12,810           &      -         &     -             & 5,781             \\ \hline
\end{tabular}%
}
\caption{The class distribution of several publicly available datasets of offensive tweets ~\cite{founta2018large,davidson2017automated,waseem:2016:NLPandCSS}. Founta et al. is the dataset selected for this data challenge. We merge the sexism (3,769) and racism (2012) as \textit{hateful} for Waseem et al.'s work.}
\label{tab:data_distri}
\end{table}

From Table~\ref{tab:data_distri}, it is easy to observe that the data distribution across the classes is severely imbalanced, especially for the \textit{hateful} class. This is a common phenomenon that is present even in large-scale datasets. This is because the \textit{normal} tweets greatly dominate the Twitter landscape, making the \textit{hateful} ones rare (not because of the lack of volume, but because they get diluted in the sea of tweets). Because of the sparse nature of hateful tweets, it is impractical for humans to manually identify the offensive tweets from a huge pool of normal ones, which seriously limits the scale of the usable labelled data. As a result, it is common for text classifiers to be severely under-trained for rare labels such as hateful speech and suffer from severe over-fitting~\cite{founta2018large}.

To solve the above two problems, we propose a powerful and easy-to-deploy data augmentation method: Dager; it uses the natural language generation (NLG) ability of current auto-regressive language models to augment source data. The idea is straightforward: we generate augmentation data samples with the same class labels as the original dataset, and add them to the training set, thus enlarging the low-resource class. In this way, we can obtain a more balanced and abundant dataset. Specifically, we first extract some semantic features from the target class, and then use these features to guide a conditional generator, thus generating the desired augmentation data. We implement our conditional generator by modifying a existing language model, GPT-2~\cite{radford2019language}, that is trained on 8 million web pages and has 1.5 billion parameters.

The advantages of Dager are three-fold: 
1) It is powerful. Our method significantly improves the performance on all test datasets trained on 4 popular classifiers. 2) It's easy to deploy. Our method does not require external datasets or rely on training auxiliary tools; instead, we make use of an off-the-shelf language model (LM) and focus on mining the potentially informative features for the low-resource data classification~\cite{liu-etal-2020-data}. 3) It has high augmentation quality. Instead of simple word-level or phrase-level replacement\cite{kobayashi2018contextual,wei2019eda}, our method provides sentence-level text data augmentation, enriching semantic and syntactic features of our generated texts. 

The goal of this work is to empirically analyze whether the current advances in large-scale language models can benefit low-resource data classification tasks. We take the offensive language detection task as an example and propose using LMs to generate augmentation data. The core module of our method is a conditional generator that can generate unlimited target label text. Its theoretical underpinning is described in the Approach section. In the Evaluation section, we systematically examine Dager in three perspectives, showing that our method provides substantial improvements on the offensive language classification task and that it is classifier-agnostic.

\section{Approach}
To enable the controlled generation of preferred attributes in an existing LM (i.e. the four class labels in this data challenge: \textit{normal}, \textit{spam}, \textit{abusive} and \textit{hateful}), we inject carefully designed conditional signal into the decoding stage. From a high-level point of view, we frame this process as a simple application of Bayesian inference, which can be presented as:

\begin{equation}
    P(x_t|c) \propto P(x_t)P(c|x_t)
\end{equation}

\noindent where $c$ is the condition signal from external input and $P(x_t|c)$ is the conditional generation output. Through basic Bayesian rule, we convert the expected posterior probability to the product of two known probabilities: 1) the prior probability $P(x_t)$, the probability we generate $x_t$ at step $t$, and 2) the conditional probability $P(c|x_t)$, which is the probability of observing condition signal $c$ given current output $x_t$. Our work focuses on how to compute the above two probabilities through simple modification on existing LMs. We choose one of the most powerful auto-regressive LMs, GPT-2~\cite{radford2019language}, as our base model, and make some necessary modifications to convert it into a conditional generator (shown in Figure~\ref{fig:conditional_generator}). The computation of the above two probabilities are shown as follows.

\subsection{Language Modeling Probability: $P(x_t)$}

Given a sequence of tokens $x_{<t} = \{x_0, x_1, x_2, ... x_{t-1}\}$ and accumulated hidden states $h_{<t}$, a vanilla auto-regressive language model is trained to maximize the probability of the next step token $\hat{x}_t$; with such design, the model will automatically pick the token with the highest probability $x_{t}$ as the $t$ step decoding output:

\begin{equation}
    x_t \sim \argmax_{\hat{x}_t} {p(\hat{x}_{t} | x_{<t})} = \textrm{LM}(x_{<t}, h_{<t})
\end{equation}

\begin{figure*}[t]
  \centering
   \includegraphics[width=0.95\textwidth]{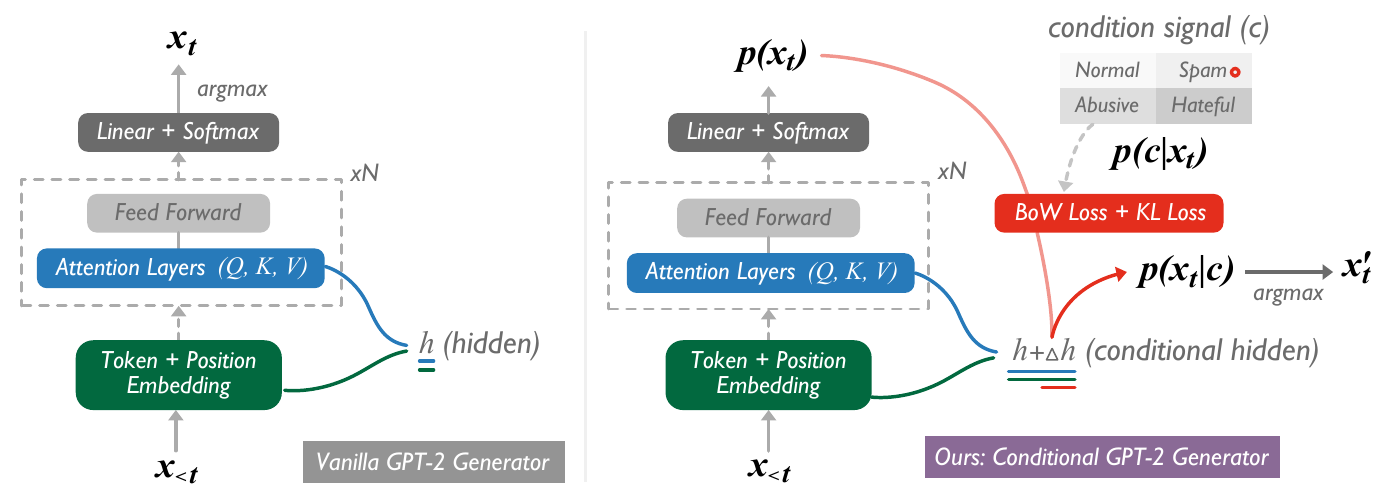}
    \caption{Overview of our conditional generator.}
   \label{fig:conditional_generator}
 \end{figure*}

In our case, however, because of the non-differentiable nature of the argmax function, we cannot perform condition signal injection after we finish the decoding. Instead, as shown in Figure~\ref{fig:conditional_generator}, we postpone the argmax decoding to a later stage and leave the space for our condition controlling procedure (described in later subsection). We also configure the LM to output the hidden states rather than the decoded tokens. The new generation is based on the cached past hidden states. The language modeling probability $P(x_t)$ can thus be computed by the softmax output of the last hidden layer, which is:

\begin{equation}
    P(x_t) = \textrm{softmax}(h_{t}), \textrm{where}\ h_{t} \sim \textrm{LM}(x_{<t}, h_{<t})
\end{equation}

\noindent Compared with the vanilla decoding manner, we use the intermediate output of the LM at step $t$ as a reasonable estimation of the token distribution, and therefore we keep all the probability computation within the space that is able to be optimized by gradients. Note that the current output only reflects the unconditional generation probability. Such a prior probability plays as a semantic limit over the whole conditional generation, which guarantees the generation quality.

\subsection{Condition Controlling Probability: $P(c|x_t)$}
\label{sec:cond}

The computation of condition controlling probability is where we inject the condition signal $c$ into the decoding stage. Given the current step generation $x_t$, $P(c|x_t)$ measures the probability of it being the output of condition signal $c$. We interpret this probability as the similarity of $x_{t}$ and the lexicon features we find in the target class and denote it as the BoW (Bag-of-Words) loss. We also consider the Kullback–Leibler (KL) divergence between the conditional and unconditional distribution of the tokens as an auxiliary balancing loss. As shown in Figure~\ref{fig:conditional_generator}, we then add an update term to the cached hidden states to force the next step generation to forward the condition signal direction. In this way, we control the gradient updates in terms of the condition signal by leveraging the existing gradient optimization path of the vanilla GPT-2 model. The add-on term $\Delta h$ is the extra gradient computed from the following two losses.

\noindent \textbf{BoW Loss.} The lexicon feature mining is through the standard TF-IDF frequency-based scoring mechanism. We collect the top 500 score tokens for each class as the feature lexicon. The BoW loss is computed as:

\begin{equation}
  \mathcal{L}_{\textrm{BoW}} = \sum_{w_i \in w_c} -\log(h_t * \textrm{Embedding}(w_i))
\end{equation}

\noindent where $w_c$ is the feature lexicon set extracted for a particular class. We compute the sum of negative log-likelihood of $x_t$ and each feature word $w_i$ in the $w_c$ set.

\noindent \textbf{KL Loss.} Although we have the prior probability $P(x_t)$ to guarantee the overall readability of the generation, we find that during the condition controlling procedure the condition controlled hidden ($h+\triangle h$) may drift too much to reach a readable generation. Therefore, we incorporate a KL divergence term to compensate the controlling drift. Specifically, the KL loss is computed over the unconditional hidden states and the conditional hidden states:

\begin{equation}
  \mathcal{L}_{\textrm{KL}} = \sum_{i=1}^t \textrm{softmax}(h_i) \cdot \log \frac{\textrm{softmax}(h_i)}{\textrm{softmax}(h_i+\triangle h)}
\end{equation}

\noindent where the conditional update term is computed as ($s$ is selected step size):

\begin{equation}
  \triangle h = -s \cdot \frac{\partial (\alpha \mathcal{L}_{\textrm{BoW}} + \beta \mathcal{L}_{\textrm{KL}})}{\partial h_i}
\end{equation}

The summation of the above two losses will be backward through the LM. Hyperparameters $\alpha$ and $\beta$ are two hyper parameters controlling the weight of each kind of loss. Our empirical results show that $\alpha = 0.3$ and $\beta = 0.01$ can achieve  high-quality conditional generation.


\section{Data Processing}

We remove the punctuation, stop words, hashtags and urls in the tweets of the data challenge dataset. We also filter out tweets whose length is above 30 tokens. After our pre-processing, we are left with 99,603 tweets in total. We further split the data into training and test set by a 4:1 ratio (training takes 80\% and test 20\%). We make sure to keep the original class distribution on the training and test set. In the following section, we run experiments on a down-sampled set of the training data, with sizes {1\%, 5\%, 20\%, 40\%, 60\%} of the whole dataset.

\section{Evaluation}
In our evaluation we attempt to answer three overarching questions:

\subsection{Does Dager Improve Performance?}

To verify the effectiveness of Dager, we set up several data starvation experiments. The experiments involve using the same evaluation dataset (20\% of the whole dataset) but different sizes of training data. From the initial training data size of 80\%, we gradually decrease the training data size to only 1\%. We then fill the down-sampled training sets with boosting samples until their size reaches 80\% of the whole data (to compare directly with the original 80\% training set).

\begin{figure}[H]
  \centering
  \includegraphics[width=0.9\linewidth]{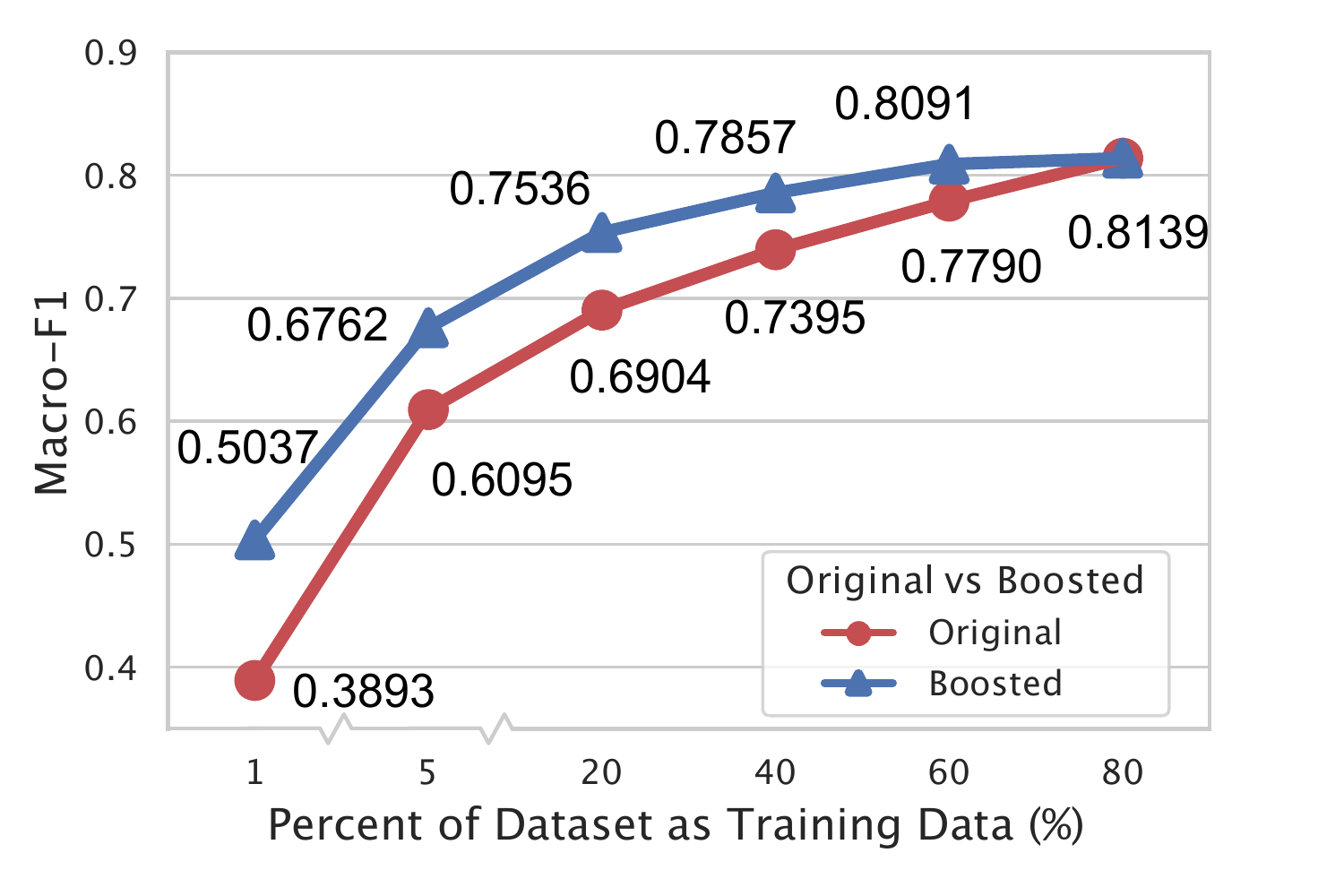}
  \caption{Macro F1 scores for different sizes of the training data, ranging from 1\% to 80\% (the remaining 20\% is used for evaluation) of the whole dataset, and the corresponding boosted training set. All the test results are averaged over five times repeat tests and the evaluation data is kept the same for all settings.}
  \label{fig:boost_works}
\end{figure}

We use the popular BERT~\cite{devlin-etal-2019-bert} sequence classifier as the judgement classifier. The red line in Figure~\ref{fig:boost_works} shows the detection performance without boosting samples: the Marco F1 score decreases from 0.8139 (when trained on the full 80\%) to 0.3893 (when trained on 1\%). The blue line shows the performance of the boosted classifiers (boosted back to 80\% using data generated by Dager), demonstrating substantial F1 improvement up to 11\% in absolute terms.

\subsection{Does Dager Preserve Class Labels?}


We are also interested in whether the generated data preserves the original true labels. We run a set of original-boosting ratio controlled experiments by gradually increasing the percentage of boosting data in the training dataset. We examine whether fusing generated samples with the original data leads to performance deterioration. At each step, we increase the amount of boosting data by 10\% of the original data size, until of the 80\% of the data used for training, 70\% is made up of Dager generated data. The results of the experiments are shown in Table~\ref{tab:boosted_same}.

\begin{table}[]
\centering
\scriptsize
\resizebox{0.47\textwidth}{!}{%
\begin{tabular}{l|cccc} \hline
\textbf{Ratio}    & 80\% / 0 (\textbf{ref})         & 70\% / 10\%      & 60\% / 20\%      & 50\% / 30\%      \\ \hline
\textbf{Original} & 79,682            & 69,721            & 59,761            & 49,801            \\
\textbf{Boosting} & 0                & 9,961             & 19,921            & 29,881            \\
\textbf{F1}       & 0.8139           & 0.8157           & 0.8091           & 0.7937           \\
$\Delta$\textbf{F1} (\%)     & 0                & $\uparrow$ 0.22\%  & $\downarrow$ 0.59\% & $\downarrow$ 2.48\% \\ \hline
\textbf{Ratio}    & 40\% / 40\%      & 30\% / 50\%      & 20\% / 60\%      & 10\% / 70\%      \\ \hline
\textbf{Original} & 39,840            & 29,880            & 19,920            & 9,960             \\
\textbf{Boosting} & 39,842            & 49,802            & 59,762            & 69,722            \\
\textbf{F1}       & 0.7857           & 0.7699           & 0.7536           & 0.7123           \\
$\Delta$\textbf{F1} (\%)     & $\downarrow$ 3.46\% & $\downarrow$ 5.41\% & $\downarrow$ 7.41\% & $\downarrow$ 12.5\%
\end{tabular}%
}
\caption{Verification tests to confirm the label preservation of generated boosting data. We keep the whole training data size the same, but control the ratio of original/boosting data. We also list the ratio and its corresponding number of samples. The performance deterioration is revealed in $\Delta$F1.}
\label{tab:boosted_same}
\end{table}

The results show that even in the extreme case, where the 80\% training set is composed of 10\% original data augmented with 70\% generated samples, we only witness a 12.5\% decrease in F1. When the fusion is half-half (40\% and 40\%), the F1 deterioration is less than 5\%. This indicates that the augmentation data generated by Dager is of good quality and preserves class labels. 

\subsection{Is Dager Classifier-Agnostic?}

\begin{table}[H]
\scriptsize
\centering
\resizebox{0.47\textwidth}{!}{%
\begin{tabular}{c|cccc}
\textbf{Training / Data (\%)} & 20\%   & 30\%   & 40\%    & 80\% (\textbf{ref})   \\ \hline
\textbf{CNN}                   & 0.668 & 0.678 & 0.744  & 0.785 \\
\textit{\textbf{+ Dager}} & 0.711 & 0.724 & 0.767  & -      \\ \hline
\textbf{RNN + Attn}            & 0.696 & 0.688 & 0.744  & 0.788 \\
\textit{\textbf{+ Dager}} & 0.764 & 0.752 & 0.778  & -      \\ \hline
\textbf{Transformer}           & 0.693 & 0.725 & 0.754  & 0.794 \\
\textit{\textbf{+ Dager}} & 0.740 & 0.745 & 0.781  & -      \\ \hline
\textbf{BERT}                  & 0.716 & 0.735 & 0.757 & 0.814 \\
\textit{\textbf{+ Dager}} & 0.720 & 0.755 & 0.784  & -     
\end{tabular}%
}
\caption{Our data augmentation performance on four different classifiers. We show the results before and after we apply Dager (which doubles the size of the data).}
\label{tab:classifier_agnostic}
\end{table}

Thus far we have verified that Dager can boost the performance of the BERT classifier, but what about other classifiers? In other words, is Dager classifier-agnostic? In addition to BERT, we pick three popular classifiers: vanilla CNN classifier, Bi-LSTM with attention mechanism, and self-attention based Transformer network~\cite{vaswani2017attention}. As shown in Table \ref{tab:classifier_agnostic}, Dager generally improves the performance of all the classifiers (from 1\% to 7\%, absolute), no matter the underlying architecture. We find Dager is especially helpful for less complicated classifiers (like CNN) in the extreme data-hungry case (given 20\% of the original data as training data). Nevertheless, Dager also benefits fine-tuning of LM-based classifiers, like BERT, even though they are very complex and are pre-trained on a large corpus.

\section{Conclusions}

In this work we present a generation-based data augmentation method using GPT-2 that can boost the performance of low-resource and imbalance data classification tasks. We demonstrate the efficacy of our method on the offensive language detection task (ICWSM-2020 data challenge task 2). We show improvements in the performance of several off-the-shelf classifier before and after we apply our data augmentation method. A future avenue for research is to experiment with the effectiveness of our augmentation method on other similar tasks.
\bibliographystyle{aaai}
\bibliography{icwsm_dc}

\end{document}